\def\year{2020}\relax
\documentclass[letterpaper]{article} 
\usepackage{aaai20}  
\usepackage{times}  
\usepackage{helvet} 
\usepackage{courier}  
\usepackage[hyphens]{url}  
\usepackage{graphicx} 
\usepackage{graphicx}  
\usepackage{multirow}
\usepackage{multicol}
\usepackage{comment}
\usepackage{amsmath}
\usepackage{amssymb}
\usepackage{xcolor}
\usepackage{subfig}

\title{On the Robustness of Face Recognition Algorithms Against Attacks and Bias}
\begingroup
\centering{
\author{Richa Singh$^{1,2}$, Akshay Agarwal$^{1}$, Maneet Singh$^1$, Shruti Nagpal$^1$, Mayank Vatsa$^{1,2}$\\
$^1$IIIT-Delhi, India; $^2$IIT-Jodhpur, India\\
{\tt\small \{rsingh, akshaya, maneets, shrutin, mayank\}@iiitd.ac.in} \\
}}
 \endgroup
 
\begin{document}

\maketitle

\begin{abstract}


Face recognition algorithms have demonstrated very high recognition performance, suggesting suitability for real world applications. Despite the enhanced accuracies, robustness of these algorithms against \textit{attacks} and \textit{bias} has been challenged. This paper summarizes different ways in which the robustness of a face recognition algorithm is challenged, which can severely affect its intended working. Different types of attacks such as physical presentation attacks, disguise/makeup, digital adversarial attacks, and morphing/tampering using GANs have been discussed. We also present a discussion on the effect of bias on face recognition models and showcase that factors such as age and gender variations affect the performance of modern algorithms. The paper also presents the potential reasons for these challenges and some of the future research directions for increasing the robustness of face recognition models.

\end{abstract}

\section{Introduction}

Face is one of the most commonly used and widely explored biometric modality for person authentication. 
Recent advances in machine learning, especially deep learning, coupled with the availability of sophisticated hardware and abundant data, have led to the development of several face recognition algorithms achieving superlative performance \cite{parkhi2015deep,amos2016openface}. Due to the advancements, automated face recognition systems are now utilized in several real world scenarios ranging from photo tagging in social media, photo organization in mobile devices to critical law enforcement applications of missing person search and suspect identification. While recognition accuracy is one of the key metrics to evaluate the effectiveness of a face recognition model, its robustness with respect to different types of data variations must also be evaluated.

\begin{figure}
    \centering
    \includegraphics[width=0.97\linewidth]{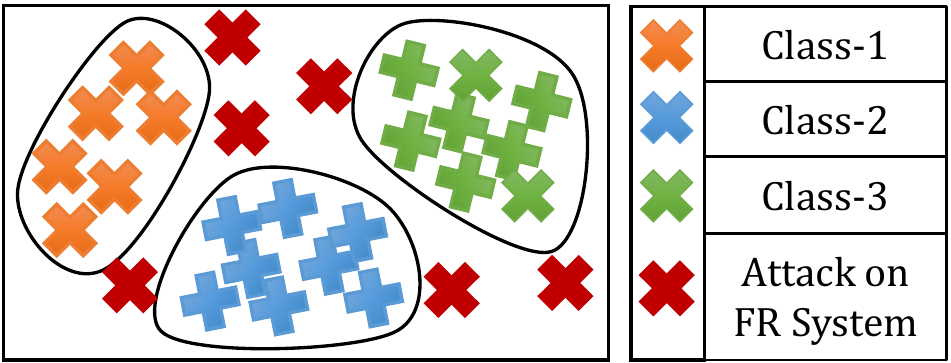}
    \caption{{An illustration where a face recognition algorithm learns the decision boundary to classify/identify three classes/subjects. The singularities between the boundaries give rise to the vulnerable points.}} 
    \label{fig:intro}
\end{figure}

Robustness of a face recognition algorithm refers to its ability to handle intentional and unintentional variations in the input space. Figure \ref{fig:intro} presents illustrative class boundaries learned by a face recognition model for three classes. The data points which are in-between boundaries (i.e. maroon crosses) can potentially be used to challenge the robustness of the algorithm (intentionally or unintentionally). Intentional variations refer to the samples/variations introduced by an \textit{adversary} which attempts to attack a face recognition algorithm for obtaining unauthorized access. For example, robustness of a recognition algorithm can be affected by different kinds of impersonation techniques such as spoofing or disguise variations. On the other hand, unintentional variations refer to the changes brought to the input image without the intent of \textit{fooling} the system. For example, variations due to unintended occlusion, disease correcting facial plastic surgery, and images captured across different characteristics such as ethnicity and gender.  

This research presents an overview of different techniques which challenge the robustness of a face recognition algorithm, along with the potential solutions provided in the literature. As part of this research, the variations observed by a face recognition system have broadly been categorized as: (i) robustness due to external adversary (i.e. attacking face recognition algorithms) and (ii) robustness with respect to bias. As mentioned earlier, an adversary can create data variations to fool the face recognition system and this can be accomplished either via (a) physical attacks or (b) digital attacks. Physical attacks refer to the techniques where changes are made to the physical appearance of a face before capturing an image. Presentation attacks, variations due to disguise/make-up, and intentional plastic surgery are some of the key techniques for physical attacks. Digital attacks refer to the changes made in the captured face image, which can result in a different output from a face recognition system as compared to the original image. For example, adversarial attacks such as the Universal perturbation \cite{moosavi2017universal} or the $l_2$-attack \cite{carlini2017towards}, as well as the variations brought due to morphing/re-touching/tampering \cite{yuan2019adversarial,scherhag2019face}. Finally, robustness with respect to ``bias'' has also been studied in this research. Biased behavior of models is a relatively recent area of research which has garnered substantial attention due to its widespread impact in the society. The inability of a recognition model to perform well for a particular subset of the population has caused concern in the community. Therefore, there is a need for an in-depth understanding of the biased behavior shown by the face recognition algorithms. 

The remainder of this paper is organized as follows: the next section elaborates upon the physical attacks, followed by a section on the digital attacks. A discussion on the robustness of face recognition models with respect to different biases are discussed thereafter, followed by the path forward for future research.

\section{Robustness Against Physical Attacks}



\begin{figure}[!t]
    \centering
    \includegraphics[width=1\linewidth]{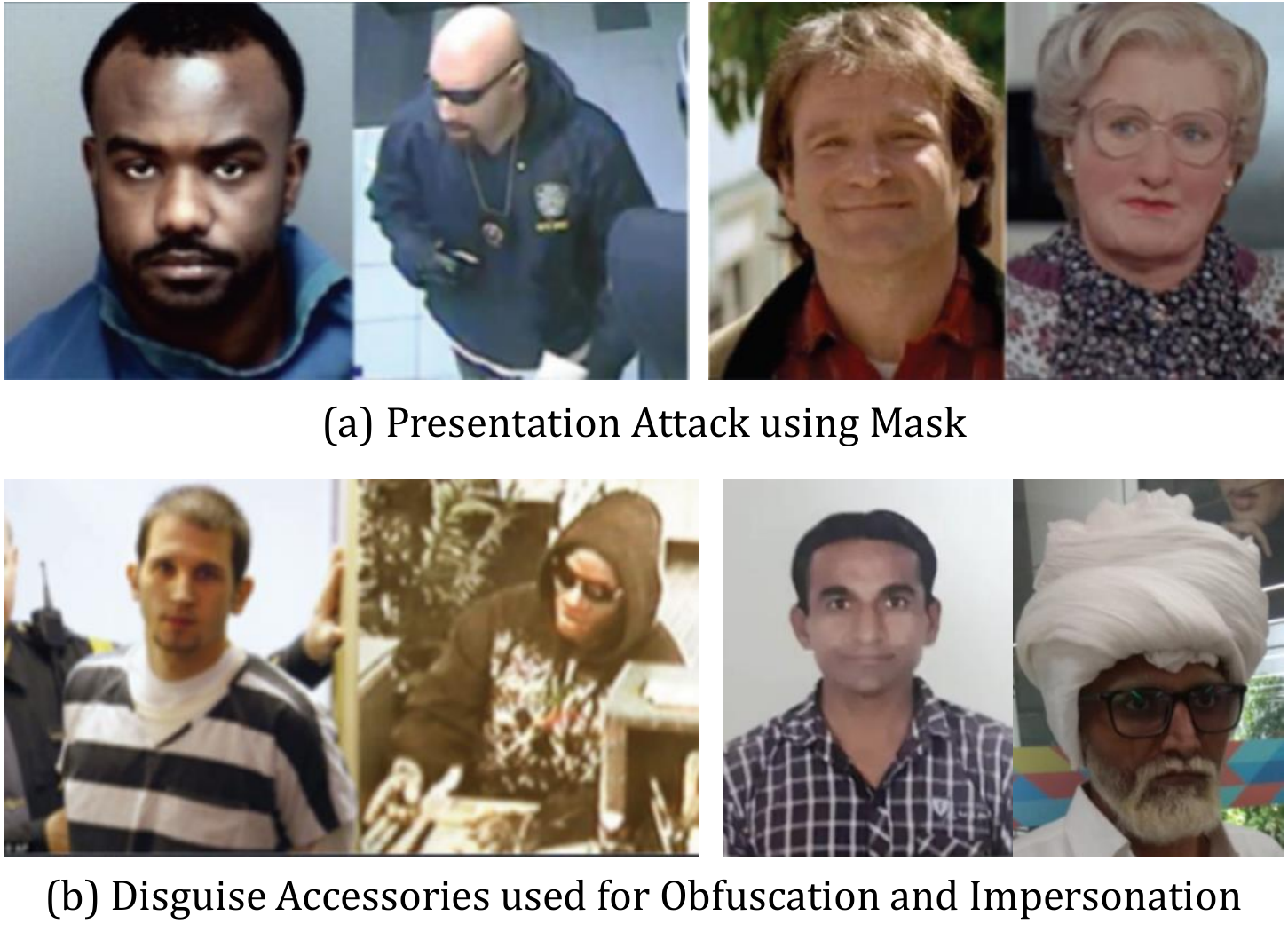}
    \caption{{Face recognition systems are susceptible to physical attacks, where physical modifications are made to the face, such as (a) presentation attack and (b) variations due to disguise accessories.}}
    \label{fig:physical}
\end{figure}

Physical attacks refer to the variations brought to the physical-self before capturing the input data for a face recognition system. In terms of a face recognition pipeline, modifications are performed at the sensor level, such that a modified face image is acquired by the recognition system. A face recognition system can be attacked with the intention of impersonating another individual (in order to obtain unauthorized access) or obfuscating one's identity. As demonstrated in Figure \ref{fig:physical}, generally physical attacks can correspond to: (i) presentation attacks, (ii) disguised faces, and (iii) variations due to plastic surgery. The following subsections elaborate upon these physical attacks in terms of the associated literature and implications.  

\subsection{Presentation Attacks}


According to the ISO/IEC JTC1 SC37 Biometrics
2016 Standards, presentation attack (Figure \ref{fig:physical}(a)) can be defined as ``an alteration in the face acquisition system with the intention of modifying its intended working''. As mentioned earlier, the aim of presentation attacks can be two fold: (i) impersonation: where an attacker wants to acquire the identity of someone else for illegal access and (ii) obfuscation: where the person wants to hide his/her own identity. In the literature, Marcel, Nixon, and Li (\citeyear{EURECOM+5667}) showed that face recognition systems are vulnerable against various presentation attacks ranging from cost-effective 2D photo mediums to sophisticated 3D silicone masks. The first public 2D photo print-attack database, namely the NUAA photo imposter (PI) database, was released in 2010. Later, Print-Attack \cite{Anjos_IJCB_2011}, CASIA-FASD \cite{zhang2012face}, and Replay-Attack \cite{Chingovska_BIOSIG-2012} databases were developed to showcase the challenging nature of these physical attacks. 
The above-listed databases, focusing only on 2D presentation mediums, generally suffer from texture loss, image quality, and Moir{\'e} patterns. To overcome these limitations and with the advancement in 3D technology, 3D masks are deployed to attack the face recognition system. Both commercial and deep learning based face recognition systems are found vulnerable towards such presentation attacks. In 2017, Manjani et al. (\citeyear{manjani2017detecting}) prepared the first silicone mask attack database containing videos captured in unconstrained settings using various YouTube links. In total, the database consists of $130$ videos of the real face and people wearing a silicone mask. Later, several other databases have been prepared showcasing the challenging nature of 3D mask attacks.  

In the literature, several image features based algorithms have been developed to detect presentation attacks \cite{galbally2014biometric,ramachandra2017presentation}. The image features based algorithms can be grouped in pre deep learning era and post deep learning success. The pre deep learning algorithms are generally based on hand-crafted features: either texture features, motion features, image quality features, or the combination of them. In the post deep learning era, architectures such as deep-dictionary learning, convolutional neural network (CNN), and long short-term memory (LSTM) have been applied for feature extraction and detection of presentation attacks. For instance, Manjani et al. (\citeyear{manjani2017detecting}) presented a novel detection algorithm using multiple level dictionary learning. The error rate of the proposed algorithm is $6$\% lower than the then best performing algorithm on the proposed silicone mask database in the seen setting. On the other hand, in open-set testing, the error rate of the proposed algorithm is at least $17$\% less than existing algorithms\footnote{Generally, PAD algorithms suffer from low generalizability, where the defense might work perfectly on a seen database and seen attack but fails under unseen settings. The scenario is popularly referred to as open-set, where during test, the defense algorithm is provided with images of unseen database or attack.}.


Face recognition systems are not limited to operate in day time under the visible spectrum; they are also deployed to operate in the infra-red spectrum (night time). Agarwal et al. (\citeyear{agarwal2017face}) presented the first-ever multi-spectral video-based face presentation attack database. Real and presentation attack videos are captured in the visible (VIS), near infra-red (NIR), and thermal spectrum. Presentation attack medium includes 3D latex masks and 2D hard resin masks of famous celebrities. Face recognition experiments performed using commercial-off-the-shelf (COTS) system and hand-crafted texture features show that the performance of the system drops when attack videos are provided for recognition. Further, to secure the system from presentation attacks, several existing texture-based feature extractors are implemented. It is found that presentation attack detection is highest in the thermal spectrum and lowest in the NIR spectrum. The best performance is obtained by the combination of multi-resolution decomposition from wavelet and texture features via a gray-level-co-occurrence matrix (GLCM) \cite{agarwal2016face}. This was followed by a novel feature aggregation based presentation attack detection algorithm for 2D and 3D attack mediums \cite{7899772}. After performing the motion magnification to enhance the micro patterns in the videos, a linear SVM classifier is trained over texture and motion features separately, followed by score-level fusion. 

Recently, new concerns have been raised regarding the robustness of face recognition systems. The PAD algorithms, which are used to protect the face recognition algorithms, are also vulnerable to attacks and unseen distribution samples. Agarwal et al. (\citeyear{agarwal2019icb}) showed that a face recognition system can be made vulnerable by tampering the feature extraction block of PAD algorithms. A convolutional autoencoder based mapping has been learned to map the features of fake class to the feature distribution of real class. Given the vulnerabilities, it is our belief that future research should focus primarily on developing (i) robust PAD algorithms and (ii) universal detectors \cite{mehtacrafting} capable of handling multiple attacks.



\subsection{Disguise and Make-up} 


Face recognition systems are often presented with the challenge of recognizing disguised faces (Figure \ref{fig:physical}(b)). Disguised face recognition is accompanied with the inherent characteristic of \textit{intent}. Disguise accessories can be used intentionally or unintentionally to obfuscate different face regions. Intentionally, disguise accessories can be used to impersonate another person in order to gain unauthorized access. Disguise accessories can also be used to obfuscate one's identity by hiding certain face parts. Similarly, usage of accessories such as sunglasses, hats, or scarves also result in unintentional disguised faces. The combination of different types of disguises, their usage, and varying intent, results in vast variations observed in disguised faces, which often tend to challenge the robustness of face recognition systems. 

The AR Face dataset \cite{ar} is one of the initial datasets containing face images with disguise variations. It contains images pertaining to 126 subjects, captured in constrained settings with a fixed set of disguise accessories (sunglasses and scarves). The dataset has served a pivotal role in promoting research on disguised face recognition; however, the constrained nature of the dataset resulted in algorithms achieving high recognition performance very quickly (almost 95\% \cite{singh09}). The AR Face dataset was superseded by other more challenging datasets, which facilitated research in the direction of disguised face recognition. In 2013, the IIITD In and Beyond Visible Spectrum Disguise database (I$^2$BVSD) \cite{dhamecha13} was released for understanding and evaluating disguised face recognition in the visible and the thermal spectra. The dataset continues to be one of the seminal datasets for evaluating face recognition systems under disguise variations. In 2014, Dhamecha et al. (\citeyear{dhamecha14Plos}) compared human performance and machine performance for the task of disguised face recognition on this dataset. The algorithm proposed by the authors identified disguised facial regions, and performed person recognition using the non-disguised facial regions only. 

Up till 2016, most of the research on disguised face recognition involved face images captured in relatively constrained settings. Wang and Kumar (\citeyear{wang16Isba}) presented a Disguised and Makeup Faces Dataset containing 2460 face images of 410 subjects. The dataset contains images collected from the Internet with variations across different disguise accessories and makeup. In 2018, the Disguised Faces in the Wild (DFW) dataset \cite{singh2019recognizing} was released as part of the International Workshop on DFW held in conjunction with CVPR2018. The DFW dataset is a first-of-its-kind dataset containing 11,157 face images of 1,000 subjects. It is the first dataset to contain multiple images for each subject, along the lines of \textit{normal}, \textit{validation}, \textit{disguised}, and \textit{impersonator}. 
Recently, the concept of the DFW dataset is further extended and the DFW2019 dataset \cite{singh19DFW} is presented as part of the DFW2019 competition at ICCV2019. This includes additional sets of plastic surgery and bridal makeup. While the top performing teams in the competition demonstrated high verification performance at higher False Acceptance Rates \cite{arcface,singh19DFW}, analysis of the submissions demonstrate low performance (less than 10\% True Acceptance Rate) at 0\% False Acceptance Rate; a metric often used in stricter settings such as access control in highly secure locations. The key observation formed in the two competitions is that impersonators are the most difficult subset. 


\subsection{Plastic Surgery} Plastic surgery is another covariate of face recognition which challenges the robustness of automated recognition systems. Plastic surgery is often performed to modify face parts such as the nose, eyes, lips, ears, or bone structure. Post surgery, an individual can demonstrate large permanent variations in the face shape or different facial regions, thereby resulting in low intra-class similarity. In 2009, plastic surgery was established as a challenging covariate for face recognition, which required dedicated research attention \cite{singh09Cvprw}. In 2010, the first publicly available plastic surgery dataset was released \cite{singh10Tifs} containing pre and post surgery images of 900 subjects. The dataset continues to be the only available dataset for this problem, which is still heavily used by the researchers across the world. 

The release of the IIITD Plastic Surgery dataset instigated the development of several face recognition algorithms capable of handling variations due to plastic surgery \cite{nappi16IVC}. Bhatt et al. (\citeyear{bhatt13Tifs}) proposed a multi-objective evolutionary granular algorithm for matching faces before and after plastic surgery. The algorithm demonstrated improved recognition performance as compared to the then state-of-the-art results. Marsico et al. (\citeyear{marsico15PR}) proposed region-based strategies for face recognition under variations due to plastic surgery. Suri et al. (\citeyear{suri18Btas}) proposed a COST framework (COlour, Shape, and Texture) for matching pre and post plastic surgery face images. For detecting faces which have undergone plastic surgery, a Multiple Projective Dictionary Learning based technique has been proposed, followed by a face verification pipeline utilizing the information from the altered and non-altered regions \cite{kohli15Access}. A high accuracy of almost 98\% is achieved in the plastic surgery detection task. Recently, the DFW2019 competition \cite{singh19DFW} has also contained a protocol for recognizing images under plastic surgery variations, where deep learning based baseline algorithms show around 50\% verification accuracy at 0.01\% False Acceptance Rate. It is our belief that the availability of these face datasets will enable deep learning based face recognition systems to encode the covariate of plastic surgery and improve the results.

\section{Robustness Against Digital Attacks}

\begin{figure}[t]
    \centering
    \includegraphics[width=0.8\linewidth]{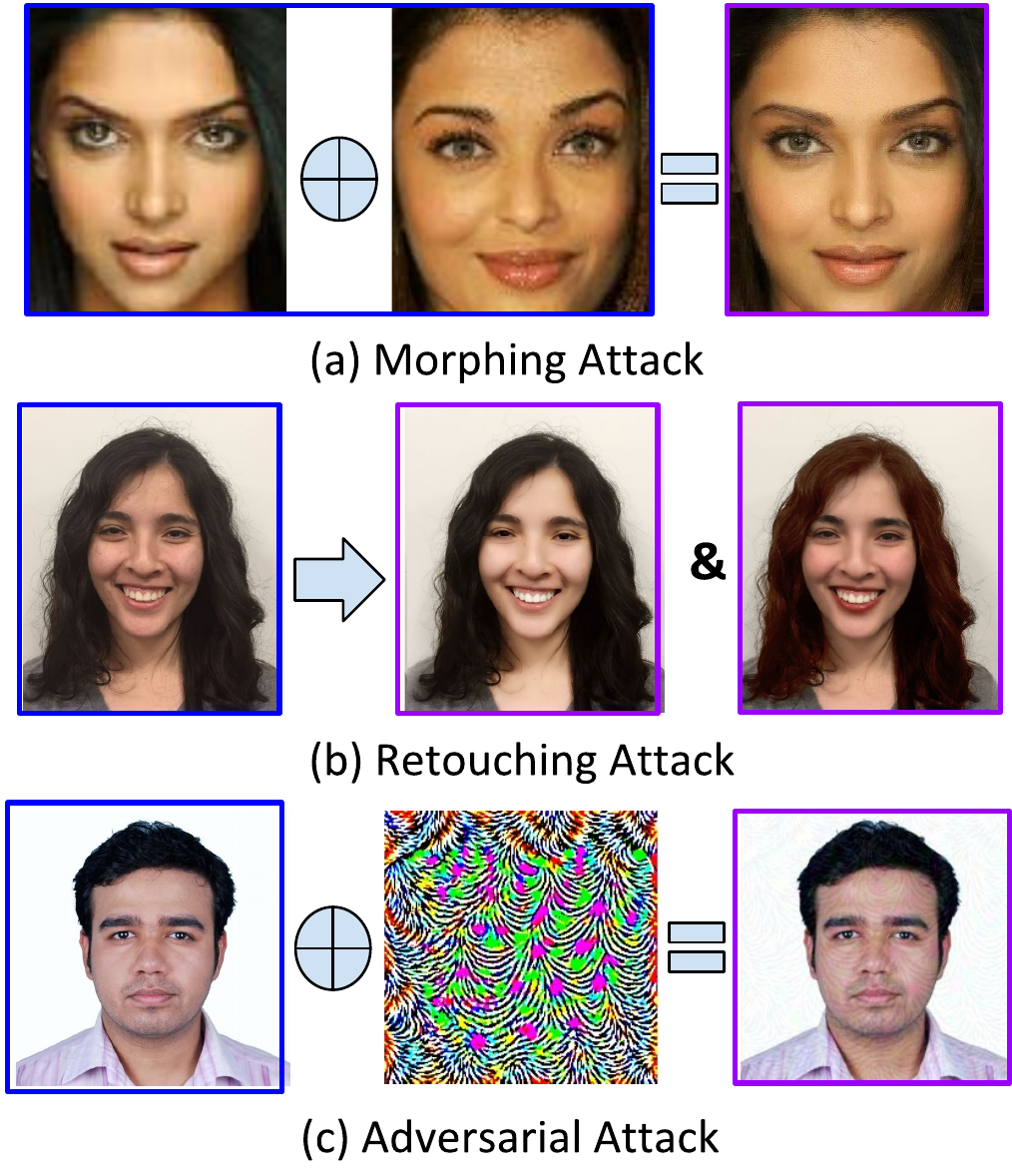}
    \caption{{Digital attacks: (a) morphing, (b) retouching, (c) adversarial perturbation. In each row, image(s) in blue box represents the real image, and remaining are attack images.}}
    \label{fig:digital}
\end{figure}

Digital attacks correspond to the variations introduced in the acquired image before presenting it to the face recognition system. With the availability of several image modification tools, it has become relatively easy for attackers to digitally modify a face image. As shown in Figure \ref{fig:digital}, generally, digital attacks can broadly be classified into: (i) adversarial attacks and (ii) alterations -  morphing/re-touching/tampering. The following subsections elaborate upon each type of digital attack and its related literature. 

\subsection{Adversarial Attacks} 


Despite the high classification performance obtained by deep learning techniques \cite{majumdar2016face,he2015delving,silver2016mastering}, they are highly susceptible to changes in the input space (Figure \ref{fig:digital}(c)). 
Szegedy et al. (\citeyear{szegedy2013intriguing}) demonstrated the vulnerabilities of CNN models by introducing a minute noise or perturbation in the input image.
Karahan et al. (\citeyear{karahan2016image}) have shown that deep face recognition algorithms are susceptible to image degradation based effects such as Gaussian noise, contrast, blur, and facial part occlusions. It is also observed that the accuracies of GoogLeNet \cite{szegedy2015going} and VGG-Face \cite{parkhi2015deep} degrade with color balance manipulation. Dabouei et al. (\citeyear{dabouei2019fast}) have perturbed the face images by manipulating various facial landmarks, and demonstrated that geometric attacks are more than $98$\% successful on the state-of-the-art face recognition networks. 


Goswami et al. (\citeyear{aaa2018goswami}) showed that several commercial and deep CNN based face recognition algorithms are vulnerable towards different adversarial attacks at (i) image-level and (ii) face-level. In the extended work \cite{ijcv2019goswami}, the authors proposed two defense algorithms: (i) adversarial perturbation detection algorithm utilizing the intermediate filter maps of a CNN, and (ii) a mitigation algorithm for recognizing adversarial faces. To mitigate the effect of adversarial noise, the most affected filter maps of a CNN model are selectively dropped out, and matching is performed using the unaffected filter maps. In another work, Agarwal et al. (\citeyear{8698548}) demonstrated that the attacks performed using image-agnostic perturbations (i.e., one noise across multiple images) can be detected using a computationally efficient algorithm based on the data distribution. Further, Goel et al. (\citeyear{8698567}) developed the first benchmark toolbox of algorithms for adversarial generation, detection, and mitigation for face recognition. Recently, Goel et al. (\citeyear{goel2019btas}) presented one of the best security mechanism, namely blockchain to protect against attacks on face recognition. Layers of CNN are converted into blocks similar to blocks in the blockchain. Each block contains the data, hash function, public and private cryptographic keys to identify any possible tampering. The proposed network is resilient to any kind of tampering including modifications to the CNN weights. While defense against adversarial samples of utmost importance, researchers have also focused on evaluating the adversarial robustness of a model \cite{carlini19Arxiv}. It is our belief that going further, researchers should focus more on understanding the cause of adversaries \cite{gilmer19ICML}, and providing robust defense mechanisms \cite{carlini18ICML}.

\subsection{Morphing, Re-touching, and Tampering}



Ferrara, Franco, and Maltoni (\citeyear{ferrara2014magic}) first  demonstrated the vulnerability of commercial face recognition systems towards morphed faces. 
Agarwal et al. (\citeyear{8272754}) generated the first video-based morphed face database using the popular social messaging application, Snapchat. The database contains videos of $129$ unique subjects. Further, the effect of face morphing is demonstrated using a commercial face recognition system and in-built iPhone face unlocking system. It is observed that both the systems are unable to protect themselves from morphed images. Recently, Majumdar et al. (\citeyear{majumdar2019evading}) performed an enhanced study on face morphing through two operations: (i) morphing two identities by blending as per a certain amount, and (ii) by partially replacing a particular part of the face from a different identity. The vulnerability of two deep face recognition algorithms, OpenFace \cite{amos2016openface} and VGG-Face, are evaluated on the tampered database. Blending and replacement of the eye region show the highest impact in the recognition performance, and both the networks demonstrate a drop of at least $30$\%. Further, to protect the integrity of these algorithms, a novel Siamese detection network is proposed which utilizes the RGB and high pass filtered images for tamper detection. Jain, Singh, and Vatsa (\citeyear{jain2018detecting}) proposed an algorithm for detecting synthetic face images generated using StarGAN \cite{choi2018stargan}. A support vector machine classifier is trained for binary classification over the softmax probabilities given by the CNN network.

Similar to morphing, facial retouching is an important application, particularly in the fashion and beauty product industry. In 2016, Bharati et al. (\citeyear{7464282}) prepared one of the most extensive facial retouching based database. The authors demonstrated that retouched face images can degrade the matching accuracy of the commercial system by up to $25$\%. A deep Boltzmann machine based model was proposed for detecting retouched images. The proposed architecture is able to perfectly detect the existing makeup based retouched images. In the follow-up work \cite{bharati2017demography}, the face retouched database was extended by covering multiple demographics regions. A novel semi-supervised network is also proposed for the detection of makeup and retouched images, which shows superior performance compared to existing algorithms.

Recently, GAN based techniques such as the FSGAN \cite{fsgan} have shown to generate seemingly real content, making it challenging even for humans to identify fake images. Moreover, the rise in DeepFakes \cite{deepfake1,deepfake2} and other sophisticated morphing techniques demands robust solutions for detection of fake content.

\section{Robustness Against Bias}
\label{sec:bias}

\begin{figure}[!t]
\centering
\subfloat[][Racial bias in existing face recognition system. ] {\includegraphics[height=0.95in]{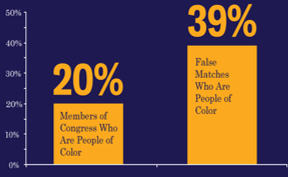}} 
\hspace{0.01em}
\subfloat[][Age bias observed in face recognition models.]{\includegraphics[height=0.95in] {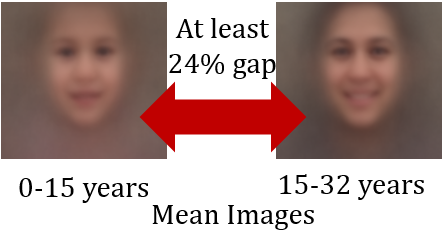}}
\caption{{(a) Recent incidents have demonstrated bias in face recognition algorithms. (b) Nagpal et al. (\citeyear{pride}) have demonstrated bias due to race and age in deep learning models.}}
\label{fig:bias}
\end{figure}

Another less explored yet crucial field for assessing robustness of face recognition systems is their invariance to the presence of \textit{bias}\footnote{As per the Oxford dictionary, bias is defined as the \textit{inclination or prejudice for or against one person or group, especially in a way considered to be unfair.}}. Recently, multiple incidents have highlighted the presence of bias in existing machine learning based systems for face analysis. Amazon's face recognition software, Rekognition, despite being easy to use, made an erroneous prediction for 28 members of the Congress and confused them with images of publicly available mug-shots. Moreover, even though only 20\% of the members of Congress are people of color, almost 40\% of the false matches belonged to them (Figure \ref{fig:bias} (a)) \cite{biasArticle}. In the literature, Buolamwini and Gebru (\citeyear{mit}) demonstrated the biased performance of three commercial software for gender classification. These algorithms performed poorly on dark skinned females as compared to lighter skinned males. The authors also introduced a new database, Pilot Parliaments Benchmark (PPB), which was labeled using the six point Fitzpatrick scale for skin color. Based on this labeled data, further analysis was performed with respect to skin tone of the subject to study the bias in existing systems. 

Following these observations, researchers have presented techniques to mitigate the effect of bias in face analysis tasks. A joint learning and unlearning framework \cite{JLU} has been proposed for eliminating bias from CNN models for age, gender, race, and pose classification from face images. A joint loss is used to optimize the network. The primary loss focuses on the task of classification, while the additional loss enforces the learnt representations to be invariant to the secondary task, and the variations in the data. Ryu, Adam, and Mitchell (\citeyear{inclusive}) proposed the \textit{Inclusive FaceNet} model, which utilized transfer learning to learn attribute prediction models for various subgroups across gender and ethnicity.
Multi-task Convolutional Neural Network (MTCNN) \cite{mtcnn} is another framework, which is proposed to learn unbiased feature representations. It jointly learns to predict the gender, ethnicity, and age from the input. Joint learning results in improved learning across sub-groups which reduces the biased behavior of the model towards a particular sub-group. Another research thread to mitigate learning biased representations involves pre-processing the data to obtain fair representations. Amini et al. (\citeyear{Amini19}) presented a pre-processing technique to de-bias face detection algorithms. The algorithm learns the latent structure of the training data with respect to the ethnicity and gender of the subject via variational autoencoders, which is later utilized to re-weight samples in order to obtain fair representations. 

Limited research has focused on understanding the effect of bias in face recognition. Recently, we have \cite{pride} presented a first-of-its-kind in-depth analysis of bias in deep learning based face recognition algorithms. We have analyzed deep learning models for the existence of bias with respect to the race and age of individuals (Figure \ref{fig:bias}(b)). It is observed that similar to humans, deep learning based face recognition models appear to undergo the phenomenon of ``own-race'' and ``own-age'' bias, where they suffer a drop in accuracy while recognizing individuals of a different race or age than those seen during training. Feature visualizations further demonstrate an inherent bias in deep learning networks, wherein they appear to focus on race-specific discriminative facial regions. These findings suggest an immediate need for researchers to focus on eliminating bias from face recognition models in order to develop \textit{fairer} systems.

\section{Discussion}

Deep learning based face recognition models may have achieved very high performance on ``seen'' distributions and learnt to predict the unseen classes under certain variations, they still show poor generalizability on unseen variations. This singularity can be exploited by an adversary to attack the models or can unintentionally yield biased decisions. For example, attacks such as adversarial perturbations, deepfakes, morphing/tampering using GANs, and silicone masks based physical presentation attacks have already been used to fool face recognition models. Future research directions should focus on two important aspects: (i) developing methods to compute the robustness level of an algorithm which assess if the algorithm would show biased behavior and (ii) developing robust defense mechanisms to build trustworthy face recognition systems. Finally, the research community will benefit from novel databases and benchmarking protocols focusing on identifying the singular points of face recognition algorithms.


\section{Acknowledgement}
The authors are partially supported through the Infosys CAI at IIIT-Delhi, India. A. Agarwal is partly supported by the Visvesvaraya PhD
Fellowship and S. Nagpal is supported via the TCS PhD fellowship. M. Vatsa is also supported through the Swarnajayanti Fellowship by the Government of India,

\small{
\bibliographystyle{aaai}
\bibliography{aaai-bib3}

\begin{thebibliography}{}

\bibitem[\protect\citeauthoryear{{Agarwal} \bgroup et al\mbox.\egroup
  }{2017a}]{8272754}
{Agarwal}, A.; {Singh}, R.; {Vatsa}, M.; and {Noore}, A.
\newblock 2017a.
\newblock Swapped! digital face presentation attack detection via weighted
  local magnitude pattern.
\newblock In {\em IEEE/IAPR IJCB},  659--665.

\bibitem[\protect\citeauthoryear{Agarwal \bgroup et al\mbox.\egroup
  }{2017b}]{agarwal2017face}
Agarwal, A.; Yadav, D.; Kohli, N.; Singh, R.; Vatsa, M.; and Noore, A.
\newblock 2017b.
\newblock Face presentation attack with latex masks in multispectral videos.
\newblock In {\em IEEE CVPRW},  81--89.

\bibitem[\protect\citeauthoryear{{Agarwal} \bgroup et al\mbox.\egroup
  }{2018}]{8698548}
{Agarwal}, A.; {Singh}, R.; {Vatsa}, M.; and {Ratha}, N.
\newblock 2018.
\newblock Are image-agnostic universal adversarial perturbations for face
  recognition difficult to detect?
\newblock In {\em IEEE BTAS},  1--7.

\bibitem[\protect\citeauthoryear{Agarwal \bgroup et al\mbox.\egroup
  }{2019}]{agarwal2019icb}
Agarwal, A.; Sehwag, A.; Vatsa, M.; and Singh, R.
\newblock 2019.
\newblock Deceiving the protector: Fooling face presentation attack detection
  algorithms.
\newblock In {\em IEEE/IAPR ICB}.

\bibitem[\protect\citeauthoryear{Agarwal, Singh, and
  Vatsa}{2016}]{agarwal2016face}
Agarwal, A.; Singh, R.; and Vatsa, M.
\newblock 2016.
\newblock Face anti-spoofing using {H}aralick features.
\newblock In {\em IEEE BTAS},  1--6.

\bibitem[\protect\citeauthoryear{Alvi, Zisserman, and Nellåker}{2018}]{JLU}
Alvi, M.; Zisserman, A.; and Nellåker, C.
\newblock 2018.
\newblock Turning a blind eye: Explicit removal of biases and variation from
  deep neural network embeddings.
\newblock In {\em ECCVW},  556--572.

\bibitem[\protect\citeauthoryear{Amerini \bgroup et al\mbox.\egroup
  }{2019}]{deepfake1}
Amerini, I.; Galteri, L.; Caldelli, R.; and Del~Bimbo, A.
\newblock 2019.
\newblock Deepfake video detection through optical flow based cnn.
\newblock In {\em IEEE ICCVW}.

\bibitem[\protect\citeauthoryear{Amini \bgroup et al\mbox.\egroup
  }{2019}]{Amini19}
Amini, A.; Soleimany, A.~P.; Schwarting, W.; Bhatia, S.~N.; and Rus, D.
\newblock 2019.
\newblock Uncovering and mitigating algorithmic bias through learned latent
  structure.
\newblock In {\em AAAI/ACM AIES},  289--295.

\bibitem[\protect\citeauthoryear{Amos \bgroup et al\mbox.\egroup
  }{2016}]{amos2016openface}
Amos, B.; Ludwiczuk, B.; Satyanarayanan, M.; et~al.
\newblock 2016.
\newblock Openface: A general-purpose face recognition library with mobile
  applications.
\newblock {\em CMU School of Computer Science} 6.

\bibitem[\protect\citeauthoryear{Anjos and Marcel}{2011}]{Anjos_IJCB_2011}
Anjos, A., and Marcel, S.
\newblock 2011.
\newblock Counter-measures to photo attacks in face recognition: a public
  database and a baseline.
\newblock In {\em IEEE IJCB},  1--7.

\bibitem[\protect\citeauthoryear{Athalye, Carlini, and
  Wagner}{2018}]{carlini18ICML}
Athalye, A.; Carlini, N.; and Wagner, D.
\newblock 2018.
\newblock Obfuscated gradients give a false sense of security: Circumventing
  defenses to adversarial examples.
\newblock In {\em ICML},  274--283.

\bibitem[\protect\citeauthoryear{{Bharati} \bgroup et al\mbox.\egroup
  }{2016}]{7464282}
{Bharati}, A.; {Singh}, R.; {Vatsa}, M.; and {Bowyer}, K.~W.
\newblock 2016.
\newblock Detecting facial retouching using supervised deep learning.
\newblock {\em IEEE TIFS} 11(9):1903--1913.

\bibitem[\protect\citeauthoryear{Bharati \bgroup et al\mbox.\egroup
  }{2017}]{bharati2017demography}
Bharati, A.; Vatsa, M.; Singh, R.; Bowyer, K.~W.; and Tong, X.
\newblock 2017.
\newblock Demography-based facial retouching detection using subclass
  supervised sparse autoencoder.
\newblock In {\em IEEE IJCB},  474--482.

\bibitem[\protect\citeauthoryear{{Bhatt} \bgroup et al\mbox.\egroup
  }{2013}]{bhatt13Tifs}
{Bhatt}, H.~S.; {Bharadwaj}, S.; {Singh}, R.; and {Vatsa}, M.
\newblock 2013.
\newblock Recognizing surgically altered face images using multiobjective
  evolutionary algorithm.
\newblock {\em IEEE TIFS} 8(1):89--100.

\bibitem[\protect\citeauthoryear{Buolamwini and Gebru}{2018}]{mit}
Buolamwini, J., and Gebru, T.
\newblock 2018.
\newblock {G}ender {S}hades: {I}ntersectional accuracy disparities in
  commercial gender classification.
\newblock In {\em ACM FAT*}, volume~81,  77--91.

\bibitem[\protect\citeauthoryear{Carlini and Wagner}{2017}]{carlini2017towards}
Carlini, N., and Wagner, D.
\newblock 2017.
\newblock Towards evaluating the robustness of neural networks.
\newblock In {\em IEEE S\&P},  39--57.

\bibitem[\protect\citeauthoryear{Carlini \bgroup et al\mbox.\egroup
  }{2019}]{carlini19Arxiv}
Carlini, N.; Athalye, A.; Papernot, N.; Brendel, W.; Rauber, J.; Tsipras, D.;
  Goodfellow, I.~J.; Madry, A.; and Kurakin, A.
\newblock 2019.
\newblock On evaluating adversarial robustness.
\newblock {\em CoRR} abs/1902.06705.

\bibitem[\protect\citeauthoryear{Chingovska, Anjos, and
  Marcel}{2012}]{Chingovska_BIOSIG-2012}
Chingovska, I.; Anjos, A.; and Marcel, S.
\newblock 2012.
\newblock On the effectiveness of local binary patterns in face anti-spoofing.
\newblock In {\em IEEE BIOSIG},  1--7.

\bibitem[\protect\citeauthoryear{Choi \bgroup et al\mbox.\egroup
  }{2018}]{choi2018stargan}
Choi, Y.; Choi, M.; Kim, M.; Ha, J.-W.; Kim, S.; and Choo, J.
\newblock 2018.
\newblock Stargan: Unified generative adversarial networks for multi-domain
  image-to-image translation.
\newblock In {\em IEEE CVPR},  8789--8797.

\bibitem[\protect\citeauthoryear{Dabouei \bgroup et al\mbox.\egroup
  }{2019}]{dabouei2019fast}
Dabouei, A.; Soleymani, S.; Dawson, J.; and Nasrabadi, N.
\newblock 2019.
\newblock Fast geometrically-perturbed adversarial faces.
\newblock In {\em IEEE WACV},  1979--1988.

\bibitem[\protect\citeauthoryear{Das, Dantcheva, and Bremond}{2018}]{mtcnn}
Das, A.; Dantcheva, A.; and Bremond, F.
\newblock 2018.
\newblock Mitigating bias in gender, age and ethnicity classification: a
  multi-task convolution neural network approach.
\newblock In {\em ECCVW},  573--585.

\bibitem[\protect\citeauthoryear{Deng and Zafeririou}{2019}]{arcface}
Deng, J., and Zafeririou, S.
\newblock 2019.
\newblock Arcface for disguised face recognition.
\newblock In {\em ICCVW}.

\bibitem[\protect\citeauthoryear{{Dhamecha} \bgroup et al\mbox.\egroup
  }{2013}]{dhamecha13}
{Dhamecha}, T.~I.; {Nigam}, A.; {Singh}, R.; and {Vatsa}, M.
\newblock 2013.
\newblock Disguise detection and face recognition in visible and thermal
  spectrums.
\newblock In {\em ICB}.

\bibitem[\protect\citeauthoryear{Dhamecha \bgroup et al\mbox.\egroup
  }{2014}]{dhamecha14Plos}
Dhamecha, T.~I.; Singh, R.; Vatsa, M.; and Kumar, A.
\newblock 2014.
\newblock Recognizing disguised faces: Human and machine evaluation.
\newblock {\em PLOS ONE} 9(7):1--16.

\bibitem[\protect\citeauthoryear{Ferrara, Franco, and
  Maltoni}{2014}]{ferrara2014magic}
Ferrara, M.; Franco, A.; and Maltoni, D.
\newblock 2014.
\newblock The magic passport.
\newblock In {\em IEEE/IAPR IJCB},  1--7.

\bibitem[\protect\citeauthoryear{Galbally, Marcel, and
  Fierrez}{2014}]{galbally2014biometric}
Galbally, J.; Marcel, S.; and Fierrez, J.
\newblock 2014.
\newblock Biometric antispoofing methods: A survey in face recognition.
\newblock {\em IEEE Access} 2:1530--1552.

\bibitem[\protect\citeauthoryear{Gilmer \bgroup et al\mbox.\egroup
  }{2019}]{gilmer19ICML}
Gilmer, J.; Ford, N.; Carlini, N.; and Cubuk, E.
\newblock 2019.
\newblock Adversarial examples are a natural consequence of test error in
  noise.
\newblock In {\em ICML},  2280--2289.

\bibitem[\protect\citeauthoryear{{Goel} \bgroup et al\mbox.\egroup
  }{2018}]{8698567}
{Goel}, A.; {Singh}, A.; {Agarwal}, A.; {Vatsa}, M.; and {Singh}, R.
\newblock 2018.
\newblock Smartbox: Benchmarking adversarial detection and mitigation
  algorithms for face recognition.
\newblock In {\em IEEE BTAS},  1--7.

\bibitem[\protect\citeauthoryear{Goel \bgroup et al\mbox.\egroup
  }{2019}]{goel2019btas}
Goel, A.; Agarwal, A.; Vatsa, M.; Singh, R.; and Ratha, N.
\newblock 2019.
\newblock Securing {CNN} model and biometric template using blockchain.
\newblock In {\em IEEE BTAS}.

\bibitem[\protect\citeauthoryear{Goswami \bgroup et al\mbox.\egroup
  }{2018}]{aaa2018goswami}
Goswami, G.; Ratha, N.; Agarwal, A.; Singh, R.; and Vatsa, M.
\newblock 2018.
\newblock Unravelling robustness of deep learning based face recognition
  against adversarial attacks.
\newblock {\em AAAI}  6829--6836.

\bibitem[\protect\citeauthoryear{Goswami \bgroup et al\mbox.\egroup
  }{2019}]{ijcv2019goswami}
Goswami, G.; Agarwal, A.; Ratha, N.; Singh, R.; and Vatsa, M.
\newblock 2019.
\newblock Detecting and mitigating adversarial perturbations for robust face
  recognition.
\newblock {\em IJCV} 127(6-7):719--742.

\bibitem[\protect\citeauthoryear{He \bgroup et al\mbox.\egroup
  }{2015}]{he2015delving}
He, K.; Zhang, X.; Ren, S.; and Sun, J.
\newblock 2015.
\newblock Delving deep into rectifiers: Surpassing human-level performance on
  imagenet classification.
\newblock In {\em IEEE ICCV},  1026--1034.

\bibitem[\protect\citeauthoryear{Jain, Singh, and
  Vatsa}{2018}]{jain2018detecting}
Jain, A.; Singh, R.; and Vatsa, M.
\newblock 2018.
\newblock On detecting {GAN}s and retouching based synthetic alterations.
\newblock In {\em IEEE BTAS},  1--7.

\bibitem[\protect\citeauthoryear{Karahan \bgroup et al\mbox.\egroup
  }{2016}]{karahan2016image}
Karahan, S.; Yildirum, M.~K.; Kirtac, K.; Rende, F.~S.; Butun, G.; and Ekenel,
  H.~K.
\newblock 2016.
\newblock How image degradations affect deep cnn-based face recognition?
\newblock In {\em IEEE BIOSIG},  1--5.

\bibitem[\protect\citeauthoryear{{Kohli}, {Yadav}, and
  {Noore}}{2015}]{kohli15Access}
{Kohli}, N.; {Yadav}, D.; and {Noore}, A.
\newblock 2015.
\newblock Multiple projective dictionary learning to detect plastic surgery for
  face verification.
\newblock {\em IEEE Access} 3:2572--2580.

\bibitem[\protect\citeauthoryear{Li and Lyu}{2019}]{deepfake2}
Li, Y., and Lyu, S.
\newblock 2019.
\newblock Exposing deepfake videos by detecting face warping artifacts.
\newblock In {\em IEEE CVPRW},  46--52.

\bibitem[\protect\citeauthoryear{Majumdar \bgroup et al\mbox.\egroup
  }{2019}]{majumdar2019evading}
Majumdar, P.; Agarwal, A.; Singh, R.; and Vatsa, M.
\newblock 2019.
\newblock Evading face recognition via partial tampering of faces.
\newblock In {\em IEEE CVPRW}.

\bibitem[\protect\citeauthoryear{Majumdar, Singh, and
  Vatsa}{2016}]{majumdar2016face}
Majumdar, A.; Singh, R.; and Vatsa, M.
\newblock 2016.
\newblock Face verification via class sparsity based supervised encoding.
\newblock {\em IEEE T-PAMI} 39(6):1273--1280.

\bibitem[\protect\citeauthoryear{Manjani \bgroup et al\mbox.\egroup
  }{2017}]{manjani2017detecting}
Manjani, I.; Tariyal, S.; Vatsa, M.; Singh, R.; and Majumdar, A.
\newblock 2017.
\newblock Detecting silicone mask-based presentation attack via deep dictionary
  learning.
\newblock {\em IEEE TIFS} 12(7):1713--1723.

\bibitem[\protect\citeauthoryear{{M}arcel \bgroup et al\mbox.\egroup
  }{2018}]{EURECOM+5667}
{M}arcel, S.; {N}ixon, M.~S.; {F}ierrez, J.; and {E}vans, N.
\newblock 2018.
\newblock {\em {H}andbook of biometric anti-spoofing : {P}resentation attack
  detection}.
\newblock {E}ditors: {M}arcel, {S}., {N}ixon, {M}.{S}., {F}ierrez, {J}.,
  {E}vans, {N}. ({E}ds.); {S}pringer {I}nternational {P}ublishing; {ISBN}:
  978-3319926261.

\bibitem[\protect\citeauthoryear{Marsico \bgroup et al\mbox.\egroup
  }{2015}]{marsico15PR}
Marsico, M.~D.; Nappi, M.; Riccio, D.; and Wechsler, H.
\newblock 2015.
\newblock Robust face recognition after plastic surgery using region-based
  approaches.
\newblock {\em PR} 48(4):1261 -- 1276.

\bibitem[\protect\citeauthoryear{Martinez and Benavente}{1998}]{ar}
Martinez, A.~M., and Benavente, R.
\newblock 1998.
\newblock The {AR} face database.
\newblock {\em CVC Technical Report}.

\bibitem[\protect\citeauthoryear{Mehta \bgroup et al\mbox.\egroup
  }{2019}]{mehtacrafting}
Mehta, S.; Uberoi, A.; Agarwal, A.; Vatsa, M.; and Singh, R.
\newblock 2019.
\newblock Crafting a panoptic face presentation attack detector.
\newblock {\em IEEE/IAPR ICB}.

\bibitem[\protect\citeauthoryear{Moosavi-Dezfooli \bgroup et al\mbox.\egroup
  }{2017}]{moosavi2017universal}
Moosavi-Dezfooli, S.-M.; Fawzi, A.; Fawzi, O.; and Frossard, P.
\newblock 2017.
\newblock Universal adversarial perturbations.
\newblock In {\em IEEE CVPR},  1765--1773.

\bibitem[\protect\citeauthoryear{Nagpal \bgroup et al\mbox.\egroup
  }{2019}]{pride}
Nagpal, S.; Singh, M.; Singh, R.; and Vatsa, M.
\newblock 2019.
\newblock Deep learning for face recognition: Pride or prejudiced?
\newblock arXiv,1904.01219.

\bibitem[\protect\citeauthoryear{Nappi, Ricciardi, and
  Tistarelli}{2016}]{nappi16IVC}
Nappi, M.; Ricciardi, S.; and Tistarelli, M.
\newblock 2016.
\newblock Deceiving faces: When plastic surgery challenges face recognition.
\newblock {\em Img. and Vis. Comp.} 54:71 -- 82.

\bibitem[\protect\citeauthoryear{Nirkin, Keller, and Hassner}{2019}]{fsgan}
Nirkin, Y.; Keller, Y.; and Hassner, T.
\newblock 2019.
\newblock {FSGAN}: {S}ubject agnostic face swapping and reenactment.
\newblock In {\em IEEE ICCV},  7184--7193.

\bibitem[\protect\citeauthoryear{Parkhi, Vedaldi, and
  Zisserman}{2015}]{parkhi2015deep}
Parkhi, O.~M.; Vedaldi, A.; and Zisserman, A.
\newblock 2015.
\newblock Deep face recognition.
\newblock In {\em {BMVC}},  41.1--41.12.

\bibitem[\protect\citeauthoryear{Ramachandra and
  Busch}{2017}]{ramachandra2017presentation}
Ramachandra, R., and Busch, C.
\newblock 2017.
\newblock Presentation attack detection methods for face recognition systems: A
  comprehensive survey.
\newblock {\em ACM Computing Surveys} 50(1):8:1--8:37.

\bibitem[\protect\citeauthoryear{Ryu, Adam, and Mitchell}{2018}]{inclusive}
Ryu, H.~J.; Adam, H.; and Mitchell, M.
\newblock 2018.
\newblock Inclusivefacenet: {I}mproving face attribute detection with race and
  gender diversity.
\newblock In {\em FAT/ML}.

\bibitem[\protect\citeauthoryear{Scherhag \bgroup et al\mbox.\egroup
  }{2019}]{scherhag2019face}
Scherhag, U.; Rathgeb, C.; Merkle, J.; Breithaupt, R.; and Busch, C.
\newblock 2019.
\newblock Face recognition systems under morphing attacks: A survey.
\newblock {\em IEEE Access} 7:23012--23026.

\bibitem[\protect\citeauthoryear{{Siddiqui} \bgroup et al\mbox.\egroup
  }{2016}]{7899772}
{Siddiqui}, T.~A.; {Bharadwaj}, S.; {Dhamecha}, T.~I.; {Agarwal}, A.; {Vatsa},
  M.; {Singh}, R.; and {Ratha}, N.
\newblock 2016.
\newblock Face anti-spoofing with multifeature videolet aggregation.
\newblock In {\em IEEE/IAPR ICPR},  1035--1040.

\bibitem[\protect\citeauthoryear{Silver \bgroup et al\mbox.\egroup
  }{2016}]{silver2016mastering}
Silver, D.; Huang, A.; Maddison, C.~J.; Guez, A.; Sifre, L.; Van Den~Driessche,
  G.; Schrittwieser, J.; Antonoglou, I.; Panneershelvam, V.; Lanctot, M.;
  et~al.
\newblock 2016.
\newblock Mastering the game of go with deep neural networks and tree search.
\newblock {\em Nature} 529(7587):484.

\bibitem[\protect\citeauthoryear{{Singh} \bgroup et al\mbox.\egroup
  }{2010}]{singh10Tifs}
{Singh}, R.; {Vatsa}, M.; {Bhatt}, H.~S.; {Bharadwaj}, S.; {Noore}, A.; and
  {Nooreyezdan}, S.~S.
\newblock 2010.
\newblock Plastic surgery: A new dimension to face recognition.
\newblock {\em IEEE TIFS} 5(3):441--448.

\bibitem[\protect\citeauthoryear{Singh \bgroup et al\mbox.\egroup
  }{2019a}]{singh19DFW}
Singh, M.; Chawla, M.; Singh, R.; Vatsa, M.; and Chellappa, R.
\newblock 2019a.
\newblock Disguised faces in the wild 2019.
\newblock In {\em IEEE ICCVW}.

\bibitem[\protect\citeauthoryear{Singh \bgroup et al\mbox.\egroup
  }{2019b}]{singh2019recognizing}
Singh, M.; Singh, R.; Vatsa, M.; Ratha, N.~K.; and Chellappa, R.
\newblock 2019b.
\newblock Recognizing disguised faces in the wild.
\newblock {\em IEEE T-BIOM} 1(2):97--108.

\bibitem[\protect\citeauthoryear{{Singh}, {Vatsa}, and
  {Noore}}{2009a}]{singh09Cvprw}
{Singh}, R.; {Vatsa}, M.; and {Noore}, A.
\newblock 2009a.
\newblock Effect of plastic surgery on face recognition: A preliminary study.
\newblock In {\em IEEE CVPRW},  72--77.

\bibitem[\protect\citeauthoryear{Singh, Vatsa, and Noore}{2009b}]{singh09}
Singh, R.; Vatsa, M.; and Noore, A.
\newblock 2009b.
\newblock Face recognition with disguise and single gallery images.
\newblock {\em Img. and Vis. Comp.} 27(3):245 -- 257.

\bibitem[\protect\citeauthoryear{Suri \bgroup et al\mbox.\egroup
  }{2018}]{suri18Btas}
Suri, S.; Sankaran, A.; Vatsa, M.; and Singh, R.
\newblock 2018.
\newblock On matching faces with alterations due to plastic surgery and
  disguise.
\newblock In {\em IEEE BTAS},  1--7.

\bibitem[\protect\citeauthoryear{Szegedy \bgroup et al\mbox.\egroup
  }{2014}]{szegedy2013intriguing}
Szegedy, C.; Zaremba, W.; Sutskever, I.; Bruna, J.; Erhan, D.; Goodfellow, I.;
  and Fergus, R.
\newblock 2014.
\newblock Intriguing properties of neural networks.
\newblock {\em ICLR}.

\bibitem[\protect\citeauthoryear{Szegedy \bgroup et al\mbox.\egroup
  }{2015}]{szegedy2015going}
Szegedy, C.; Liu, W.; Jia, Y.; Sermanet, P.; Reed, S.; Anguelov, D.; Erhan, D.;
  Vanhoucke, V.; and Rabinovich, A.
\newblock 2015.
\newblock Going deeper with convolutions.
\newblock In {\em IEEE CVPR},  1--9.

\bibitem[\protect\citeauthoryear{{Wang} and {Kumar}}{2016}]{wang16Isba}
{Wang}, T.~Y., and {Kumar}, A.
\newblock 2016.
\newblock Recognizing human faces under disguise and makeup.
\newblock In {\em IEEE ISBA},  1--7.

\bibitem[\protect\citeauthoryear{Wong}{2019}]{biasArticle}
Wong, Q.
\newblock 2019.
\newblock Why facial recognition's racial bias problem is so hard to crack.
\newblock
  https://www.cnet.com/news/why-facial-recognitions-racial-bias-problem-is-so-hard-to-crack/.

\bibitem[\protect\citeauthoryear{Yuan \bgroup et al\mbox.\egroup
  }{2019}]{yuan2019adversarial}
Yuan, X.; He, P.; Zhu, Q.; and Li, X.
\newblock 2019.
\newblock Adversarial examples: Attacks and defenses for deep learning.
\newblock {\em IEEE TNNLS} 30(9):2805--2824.

\bibitem[\protect\citeauthoryear{Zhang \bgroup et al\mbox.\egroup
  }{2012}]{zhang2012face}
Zhang, Z.; Yan, J.; Liu, S.; Lei, Z.; Yi, D.; and Li, S.~Z.
\newblock 2012.
\newblock A face antispoofing database with diverse attacks.
\newblock In {\em IEEE/IAPR ICB},  26--31.

\end{thebibliography}
}

\end{document}